\documentclass{article}
\usepackage{ijcai26}

\usepackage{times}
\usepackage{soul}
\usepackage{url}
\usepackage[hidelinks]{hyperref}
\usepackage[utf8]{inputenc}
\usepackage[small]{caption}
\usepackage{graphicx}
\usepackage{amsmath}
\usepackage{amsthm}
\usepackage{booktabs}
\usepackage{algorithm}
\usepackage{algorithmic}
\usepackage[switch]{lineno}
\usepackage{booktabs}
\usepackage{colortbl}
\usepackage[table]{xcolor}  
\usepackage[authoryear]{natbib}

\title{LocationAgent: A Hierarchical Agent for Image Geolocation via Decoupling Strategy and Evidence from Parametric Knowledge}

\author{
    Qiujun Li$^{1,2}$\and
    Zijin Xiao$^{2}$\and
    Xulin Wang$^{2}$\and
    Zhidan Ma$^{2}$\and
    Cheng Yang$^{1}$\And
    Haifeng Li$^{1}$ \\ % 这里是最后一名作者的换行
    \affiliations
    $^1$Central South University\\
    $^2$ByteDance (China)\\
    \emails
    csu\_lqq@csu.edu.cn,
    \{xiaozijin, wangxulin.1217, mazhidan\}@bytedance.com,
    ychades150@gmail.com,
    lihaifeng@csu.edu.cn
}
\begin{document}

\maketitle

\begin{abstract}
    Image geolocation aims to infer capture locations based on visual content. Fundamentally, this constitutes a reasoning process composed of \textit{hypothesis-verification cycles}, requiring models to possess both geospatial reasoning capabilities and the ability to verify evidence against geographic facts. Existing methods typically internalize location knowledge and reasoning patterns into static memory via supervised training or trajectory-based reinforcement fine-tuning. Consequently, these methods are prone to factual hallucinations and generalization bottlenecks in open-world settings or scenarios requiring dynamic knowledge. To address these challenges, we propose a Hierarchical Localization Agent, called LocationAgent. Our core philosophy is to retain hierarchical reasoning logic within the model while offloading the verification of geographic evidence to external tools. To implement hierarchical reasoning, we design the RER architecture (Reasoner-Executor-Recorder), which employs role separation and context compression to prevent the drifting problem in multi-step reasoning. For evidence verification, we construct a suite of clue exploration tools that provide diverse evidence to support location reasoning. Furthermore, to address data leakage and the scarcity of Chinese data in existing datasets, we introduce CCL-Bench (China City Location Bench), an image geolocation benchmark encompassing various scene granularities and difficulty levels. Extensive experiments demonstrate that LocationAgent significantly outperforms existing methods by at least 30\% in zero-shot settings.
\end{abstract}

\section{Introduction}
Image geolocation aims to resolve the alignment between visual content and spatial coordinates. With broad applications in fields such as criminal investigation, crisis response and travel recommendations\citep{chalvatzaras2022survey,firmansyah2024empowering}.

Existing approaches can be broadly divided into two paradigms according to how this alignment is modeled: implicit modeling methods and explicit reasoning methods. Implicit methods learn mappings between visual features and locations through supervised training on large-scale labeled datasets. In contrast, explicit reasoning methods leverage large models to emulate human expert cognition by identifying key visual clues, such as buildings or text, and conducting step-by-step inference. Although both paradigms have shown progress in certain scenarios, they also exhibit clear limitations. 

Implicit modeling methods \citep{pramanick2022world,yi2025geolocsft} focus on learning discriminative representations from large scale data\citep{zhou2024img2loc}. They perform well in coarse-grained localization tasks, such as intercontinental or country level localization, but their performance degrades significantly in fine-grained settings. This limitation stems from an inherent trade-off in feature representation: coarse-grained localization favors spatial invariance, whereas fine-grained localization requires sensitivity to subtle visual differences. Moreover, implicit methods strongly depend on the coverage of the training data distribution, which restricts their generalization ability to unseen locations.

Recent advances in large multimodal models have introduced reasoning-based approaches to mitigate the feature trade-off and generalization issues of implicit methods \citep{Li2024georeansoner,li2025globe,zhang2024retrain}. However, reasoning that relies solely on parameterized memory remains insufficient. While techniques such as prompt learning and reinforcement learning can instill certain reasoning patterns, the acquired knowledge is static. Consequently, models struggle to reliably recall factual information—such as the geographic origin of a phone number displayed on a shop sign—and instead tend to produce plausible but incorrect, hallucinated locations.

To address these limitations, we advocate reformulating image geolocation as an abductive reasoning and constraint satisfaction process driven by multimodal evidence, rather than as a pure memory retrieval task. This formulation requires not only effective localization reasoning, but also explicit mechanisms for evidence verification akin to human experts. Based on this perspective, we propose LocationAgent (Hierarchical Localization Agent), which explicitly decouples location reasoning from evidence verification. To support hierarchical reasoning and mitigate drifting in multi-step agent processes, we design the RER structure, which constrains the model’s reasoning trajectories through a structured action space and maintains clear self-awareness of its reasoning state via contextual recording.

To address insufficient data coverage for the China region and mitigate potential data leakage issues in existing benchmarks, we introduce a novel localization benchmark named CCL-Bench. Unlike prior benchmarks dominated by street view imagery, CCL-Bench sources all test images from real world internet scenarios, which better reflects practical application settings. Finally, the core contributions of this work are summarized as follows:
\begin{itemize}
\item Redefining image geolocation. We formulate image geolocation as an iterative process of reasoning and constraint satisfaction guided by multimodal evidence.
\item Proposing the LocationAgent framework. We introduce the LocationAgent framework, which decouples reasoning strategies from evidence verification. The proposed RER architecture ensures stable and accurate multi-step reasoning through role separation and context control, while hierarchical location exploration tools provide verifiable geographic evidence at different spatial levels.
\item Construction and release of CCL-Bench. We introduce CCL-Bench, a benchmark designed to address the scarcity of data for the China region in existing datasets. With fine-grained annotations on scene categories and difficulty levels.

\end{itemize}

\section{Related Work}
\subsection{Image Geolocation Methods}
Existing image geolocation methods can be broadly categorized into implicit modeling–based and explicit reasoning–based approaches. 

Implicit methods optimize losses over geographic labels to learn latent image–location mappings \citep{clark2023we,jia2025georanker,haas2024pigeon,hou2025multi}, using stronger backbones, refined geocell partitioning, or contrastive learning to improve feature alignment \citep{xu2024addressclip,hu2024progeo,vivanco2023geoclip,jia2024g3,shatwell2025gt}. However, they face two fundamental limitations: a multiscale feature conflict—where coarse-grained localization requires spatial invariance while fine-grained localization demands sensitivity to visual details, and an overreliance on memorization, which limits their ability to generalize and reason about unseen scenarios.

Explicit reasoning methods aim to mimic human geographic inference by guiding multimodal large models to identify salient visual clues and perform step-by-step reasoning through techniques such as chain-of-thought prompting \citep{Li2024StreetviewLLMEG,liu2024imagecot}, instruction fine-tuning, or reinforcement learning\citep{dou2024gaga,campos2025gaea,wang2025gre}. While these approaches improve localization accuracy and generalization, they still entangle reasoning strategies with evidence verification\citep{zhang2025navig}. In practice, they attempt to internalize both reasoning procedures and geographic facts into static parameters, despite geolocation requiring extensive region-specific and fine-grained knowledge that is difficult to capture through limited retraining.

\subsection{Image Geolocation Benchmarks}
Existing image geolocation benchmarks can be broadly classified into two categories.

The first category focuses on evaluating localization accuracy \citep{hays2008im2gps,astruc2024openstreetview}, exemplified by early datasets such as Im2GPS3K and YFCC4K \citep{vo2017revisiting}. These benchmarks primarily consist of images paired with their corresponding geographic coordinates, sourced from social media platforms or street view services. However, such datasets suffer from two major limitations. First, data leakage risks: publicly available online images may have been included in MLLM training corpora, leading to inflated evaluation results \citep{wang2024llmgeo,huang2025vlms}. Second, distribution bias: street view images differ significantly from user captured photos in practical scenarios (e.g., crisis response, daily photography) in terms of perspective, composition and image quality.

The second category consists of reasoning oriented benchmarks\citep{li2025pixels,mendes2024granular}, such as GeoComp~\citep{song2025geolocation} and GeoChain \citep{yAerramilli-etal-2025-geochain}. These benchmarks annotate images with rich semantic clues (e.g., architectural styles, textual information), designed to evaluate the step-by-step reasoning capabilities of models. Such benchmarks' sources are largely restricted to street view data from developed regions. Consequently, the scarcity of data from regions such as China limits the assessment of model generalizability in global deployment scenarios.

\section{Method}
As discussed above, implicit modeling methods struggle to balance multi scale information within a unified feature space, whereas explicit modeling methods, despite enabling hierarchical reasoning, suffer from the entanglement of reasoning strategies and verification evidence. To address these limitations, we propose LocationAgent, which explicitly decouples reasoning strategies from evidence verification.

\begin{figure}[h]  % htbp：浮动体位置参数（here/top/bottom/page）
    \centering  % 图片居中
    \includegraphics[width=0.5\textwidth]{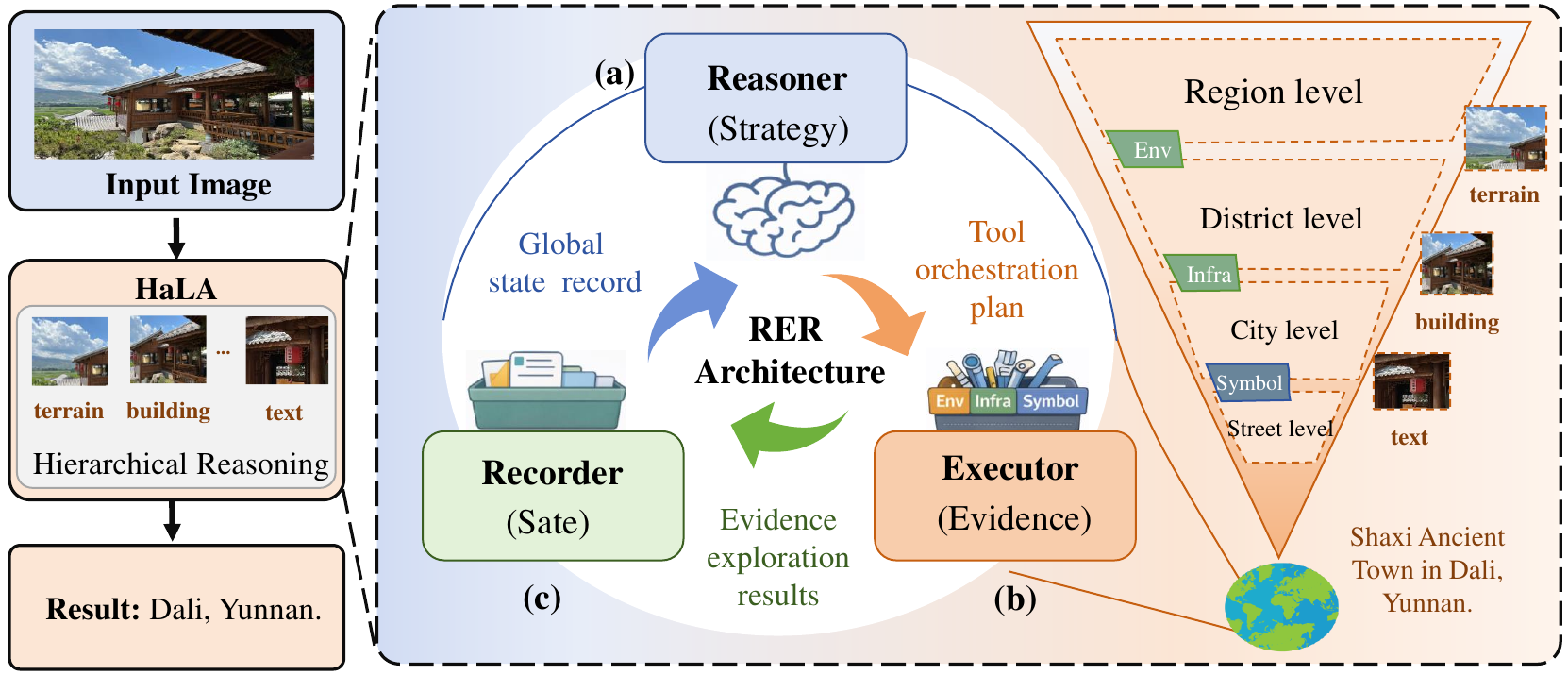}  % 核心命令：width设置宽度
    \caption{Overview of the LocationAgent framework. We reformulate image geolocation as a hierarchical abductive reasoning process. (a) The Reasoner performs strategic planning within a structured action space. (b) The Executor decouples reasoning from parametric knowledge by fetching multi-dimensional evidence from external tools. (c) The Recorder maintains state consistency to prevent reasoning drift.}  % 图片标题
    \label{fig:1}  % 图片标签（用于交叉引用）
\end{figure}

\subsection{Geolocation as an Abductive Constraint Satisfaction Process}
Let \textit{I} denote the input query image. The goal of image geolocation is to find the optimal location estimate within the continuous global coordinate space \textit{L}. Under the implicit modeling paradigm, this process is typically simplified into a single mapping function. However, such an end-to-end paradigm overlooks the logical reasoning process that is crucial to geolocation. 

As discussed above, we view geolocation as being jointly driven by reasoning strategies and evidence verification. This is analogous to a “cooking process”: reasoning strategies (cooking skills and tools) represent the model’s internal capability to process visual clues, determine exploration directions, and perform logical inference. Evidence verification (ingredients) represents geographic facts obtained from the physical world, which serve as the material basis for constructing localization conclusions. Specifically, the evolutionary logic of an image geolocation system can be summarized by the following set of formulations:
\begin{align}
a_{t+1} &= \mathcal{R}(I, \mathcal{A}_{\text{tool}}, S_{1:t}) \label{eq:reasoner} \\
e_{t+1} &= \mathcal{V}(a_{t+1}, \mathcal{W}) \label{eq:verifier} \\
L_{t+1} &= \{\, l \in L_t \mid \text{Consistent}(l, e_{t+1}) \,\} \label{eq:projection}
\end{align}

At the t reasoning step, the system state $S_t=\{L_t,E_t\}$ consists of the current candidate location region $L_t$ and the verified evidence chain $E_t$. Based on the input image I, the historical reasoning state $S_t$, and the action space $A_{tool}$, the reasoning strategy generates the next probing action to further narrow the candidate location level. Subsequently, the evidence verification module executes this action by extracting new geographic constraints $e_t$ from the external world $W$. Each newly acquired piece of evidence $e_t$ projects the current search space $L_t$ onto a smaller subset $L_{t+1}$.

\subsection{Overview of the LocationAgent Architecture}
Based on the formal definitions introduced in Section 3.1, we design the LocationAgent. As shown in Fig.\ref{fig:1}, this architecture consists of three collaborative modules: the Reasoner, which implements the reasoning strategy $R$; the Executor, which implements the evidence projection operator $V$; and the Recorder, which maintains the state set $S_t$.

\subsubsection{Reasoner: Hierarchical Spatial Reasoning Strategy}
The Reasoner serves as the cognitive core of the system. Its primary task is to determine the optimal probing action. 

The Reasoner adopts a hierarchical reasoning strategy: it first analyzes macro-level clues in the image to establish high-level spatial priors; guided by these priors, the system then gradually shifts to meso-level analyses of architectural styles and micro-level analyses of symbolic elements. Unlike existing works that rely on reinforcement learning to acquire reasoning strategies, we advocate shaping the model’s reasoning strategy by constructing a structured action space. 

We argue that general agent orchestration capabilities can be transformed into specialized geolocation reasoning strategies, provided that the action space embeds domain expert logic. In LocationAgent, the four exploration modules of the Executor perform location probing of visual elements from multiple dimensions; as a result, the Reasoner only needs to apply its general task decomposition and tool orchestration capabilities within this specialized action space. Specifically, the Reasoner’s decision making process is formulated as a sequential planning problem under a constrained action space. Supported by the hierarchical design of the action space (from macro environments to micro symbols; see next subsubsection), the module produces reasoning trajectories that align with expert intuition while maximizing information gain.

\subsubsection{Executor: Structured Action Space and Capabilities}
The Executor serves as the interaction bridge between the cognitive space and the external environment. We formalize heterogeneous knowledge sources for geolocation into four categories of core capability modules. Together, these modules constitute the system’s action space, providing channels for acquiring multimodal evidence while characterizing expert cognitive pathways from macro- to micro-levels through their functional distinctions. 

We find that accurate geolocation relies on consistent assessment of terrain and vegetation, transportation infrastructure, sociocultural symbols, and visual features. Accordingly, we design four categories of core capability modules to constitute the action space of the Executor:
\begin{itemize}
\item \textbf{Environmental Module}.This module focuses on analyzing macro-geographic characteristics, including terrain, vegetation, and climate indicators. It is designed to rapidly narrow down large scale candidate regions (e.g., North vs. South China, Southeast hilly regions) during the early stages of reasoning, thereby reducing the overall search space.
\item \textbf{Infrastructure Module}. This module specializes in extracting region specific visual clues related to infrastructure, such as architectural styles, traffic characteristics, and public facilities. It supports meso-scale localization by identifying city types, economic development levels, or specific administrative regions.
\item \textbf{Semantic Symbol Module}. This module concentrates on recognizing symbolic and textual information, including street signs, store names, zip codes, among others. By mapping these sociocultural clues to concrete administrative divisions or street-level information, the module enables fine-grained location anchoring.
\item \textbf{Image Matching Module}. Leveraging image retrieval techniques, this module performs consistency checks between extracted visual clues and a nationwide geographic database. It provides visual evidence to support reasoning conclusions and outputs candidate locations to assist decision making when the reasoning process remains ambiguous.
\end{itemize}
It is important to emphasize that although the above modules are logically organized in a hierarchical manner from macro to micro levels, this does not imply that the Reasoner must strictly follow a fixed linear invocation order. Instead, the invocation priority of modules is determined by the salience of visual clues. 

Each capability module processes a specific category of visual elements, and the same category may correspond to entirely different levels depending on the context. For example, although both are shop signs, “McDonald’s” typically serves only as a meso-level clue, indicating that the area has a certain level of commercialization. In contrast, “Bainian Yili” immediately becomes a strong micro-level clue, directly narrowing the location to Beijing and its surrounding regions, thereby allowing the model to skip macro-level environmental analysis.

\begin{figure}[h]  % htbp：浮动体位置参数（here/top/bottom/page）
    \centering  % 图片居中
    \includegraphics[width=0.4\textwidth]{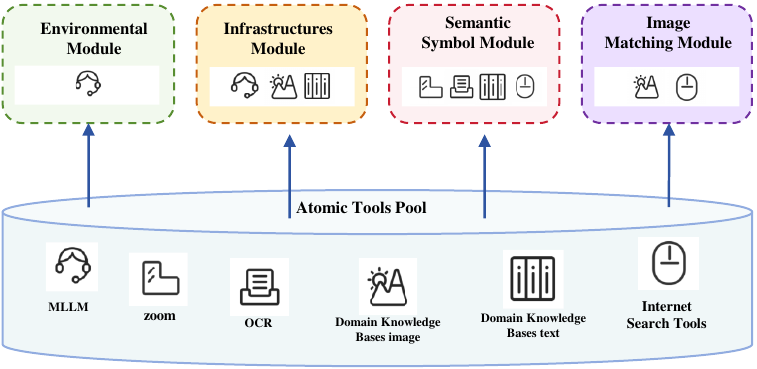}  % 核心命令：width设置宽度
    \caption{Composition relationships between atomic tools and different capability modules.}  % 图片标题
    \label{fig:2}  % 图片标签（用于交叉引用）
\end{figure}

To support high-level capability modules, LocationAgent integrates three types of atomic tools.As shown in Fig.\ref{fig:2}, the atomic tools are : perception enhancement tools (e.g., image captioning, cropping, and OCR) that transform images into multimodal semantic representations; domain specific knowledge bases that provide region aware priors for infrastructure and sociocultural analysis; and open domain retrieval tools that supply long-tail geographic information to complement the model’s internal knowledge.

\subsubsection{Recorder: State Tracking and Context Management}
The Recorder is responsible for maintaining the trajectory state of the entire reasoning process and serves as a key component for mitigating drifting in long horizon reasoning, where models lose track of their current state due to excessively long contexts, leading to issues such as repeated tool invocations and fabricated evidence. 

It dynamically updates and stores the complete interaction history from the initial step to the current time, including the executed action sequence, the set of acquired evidence, and the current candidate geographic space. By maintaining state consistency, the Recorder provides the Reasoner with coherent memory, enabling it to reflect and revise its decisions at each step based on complete contextual information, thereby improving the robustness of the reasoning trajectory.

\subsection{CCL-Bench}
\begin{figure*}[t]  % htbp：浮动体位置参数（here/top/bottom/page）
    \centering  % 图片居中
    \includegraphics[width=0.8\textwidth]{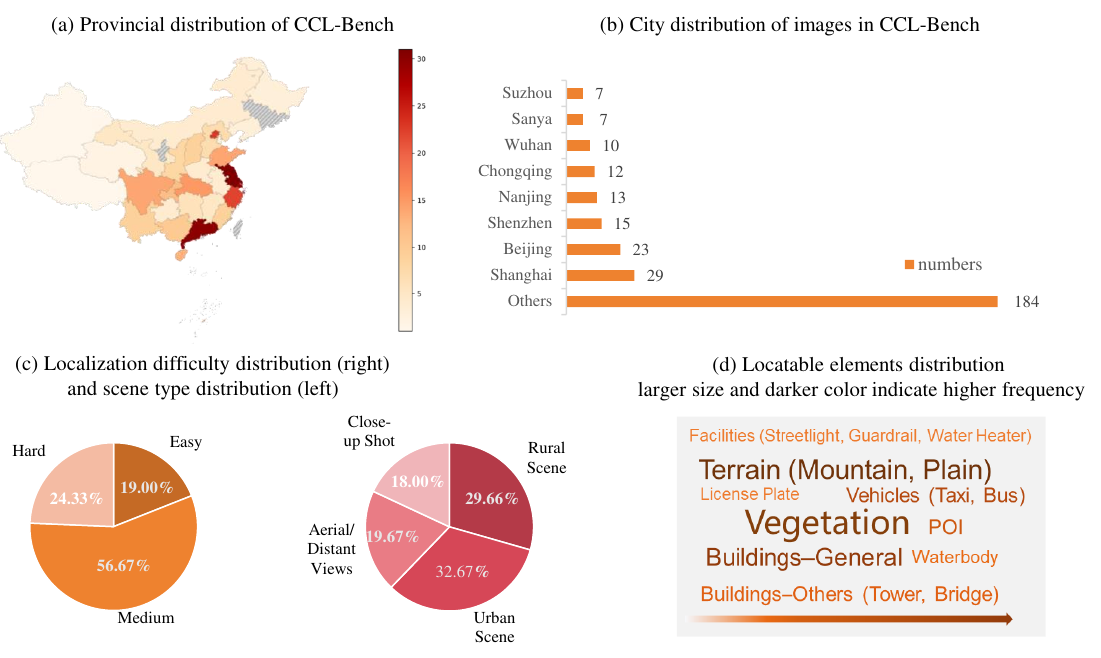}  % 核心命令：width设置宽度
    \caption{Data distribution of CCL-Bench across different categories}  % 图片标题
    \label{fig:3}  % 图片标签（用于交叉引用）
\end{figure*}
This section describes the procedures for data collection, annotation, and classification. Existing image geolocation benchmarks suffer from significant limitations in their coverage of China and in their alignment with real world image distributions. In particular, these benchmarks often fail to generalize from street-view–dominated datasets to user captured imagery. To systematically evaluate model reasoning in complex environments, we introduce CCL-Bench, a dataset designed for multimodal reasoning and image geolocation within China. Its construction is detailed from three aspects: data acquisition principles, scene distribution design, and annotation strategies.

\subsubsection{Determination of Locatability}
Instead of relying on model based prefiltering as seen in existing benchmarks, we adopt human expert solvability as the primary criterion for image locatability. 

We argue that model driven filtering introduces an evaluation bias: it tends to retain samples that align with the perceptual biases of current models while systematically excluding high difficulty cases that require long-tail knowledge or subtle visual clues. Consequently, benchmarks relying on such filtering may fail to accurately measure model performance in geographic reasoning tasks that approach human-level complexity

Specifically, candidate samples are sourced from the “Open Source Investigations” \footnote{A section of Tuxun (\url{https://tuxun.fun/interact/challenge}) where users freely upload images and players can use the web search and other tools to reason about the location of the images.} section of Tuxun. We collected real world photographs verified as solvable by human participants, ensuring that every image is demonstrably locatable through human reasoning. After selecting images from the categories, we ask each annotator to attempt geolocation. Only images that can be successfully localized by the annotators are retained.

\subsubsection{Scene Diversity}
Beyond locatability, CCL-Bench is designed to evaluate model robustness across diverse geographic and visual conditions. The benchmark encompasses four representative categories of real world scenes:
\begin{itemize}
\item \textbf{Rural Scene}. Non-urban areas with sparse visual clues, such as farmlands, low-density settlements, and natural regions, where localization relies mainly on landforms, vegetation, and subtle human traces
\item \textbf{Urban Scene}. Information-dense urban environments with dense buildings, complex road networks, and active traffic or pedestrian flow, similar to conventional street-view imagery.
\item \textbf{Aerial and Distant Views}. Macro-scale scenes captured from elevated view-points (e.g., high-rises or drones), emphasizing regional layouts, skylines, and large scale landmarks.
\item \textbf{Close-up Shot}. Scenes centered on specific buildings, ranging from full-frame landmarks to close-up views of architectural textures and decorative details.
\end{itemize}
\subsubsection{Data Annotation Scheme}
To enable interpretable analysis of model reasoning, we perform fine-grained manual annotation of visual clues that may indicate location information in CCL-Bench, including natural environmental clues, infrastructure, transportation facilities, and cultural or symbolic elements.

Based on the density and distinctiveness of available clues, CCL-Bench is further stratified into three difficulty levels: Easy, Medium, and Hard. Images containing explicit text or prominent landmarks are categorized as Easy, whereas those that rely primarily on subtle environmental clues or generic street features are classified as Hard. Overall, CCL-Bench consists of 300 high quality images with fine-grained annotations, enabling rigorous and interpretable evaluation of image geolocation models under realistic conditions.

\section{Experiments}
\subsection{Dataset Analysis}

We propose a new benchmark, with detailed information summarized in Fig.\ref{fig:3}. 
\subsubsection{Geographic Coverage }
As illustrated in Fig.\ref{fig:3} (a–b), CCL-Bench exhibits high geographic diversity, covering 30 of China’s 34 provincial level administrative regions—an 88.24\% coverage rate that rises to 93.75\% for mainland China. Consistent with the natural distribution of web sourced imagery, samples are concentrated in major urban clusters. Images from first-tier and "new first-tier" cities account for approximately 34.00\% of the dataset. Although the city level coverage is 35.84\%, CCL-Bench spans China’s primary geographic zones and cultural core areas, providing sufficient diversity for robust evaluation.

\subsubsection{Scene Type and Difficulty Distribution}
Localization difficulty in CCL-Bench follows an approximately normal distribution (Fig.\ref{fig:3} (c)), with medium difficulty samples comprising the majority (56.67\%). This distribution prevents performance saturation from overly simple cases while maintaining discriminative power by avoiding excessive difficulty. In terms of scene composition, the ratio of urban scenes to rural/natural scenes is nearly 1:1. Similarly, aerial wide angle views and architectural close-ups are balanced. This cross-scale and cross-environment distribution requires models to robustly extract visual clues across different hierarchies.

\subsubsection{Distribution of Locatable Elements }
We further analyze the visual elements supporting localization reasoning. As shown in Fig.\ref{fig:3} (d), vegetation and terrain are the most prevalent clues, followed by distinctive architectural forms and appearances. Additionally, vehicles, POIs, and other structures serve as critical references for urban localization.

\begin{figure*}[t] % htbp：浮动体位置参数（here/top/bottom/page）
    \centering  % 图片居中
    \includegraphics[width=1.0\textwidth]{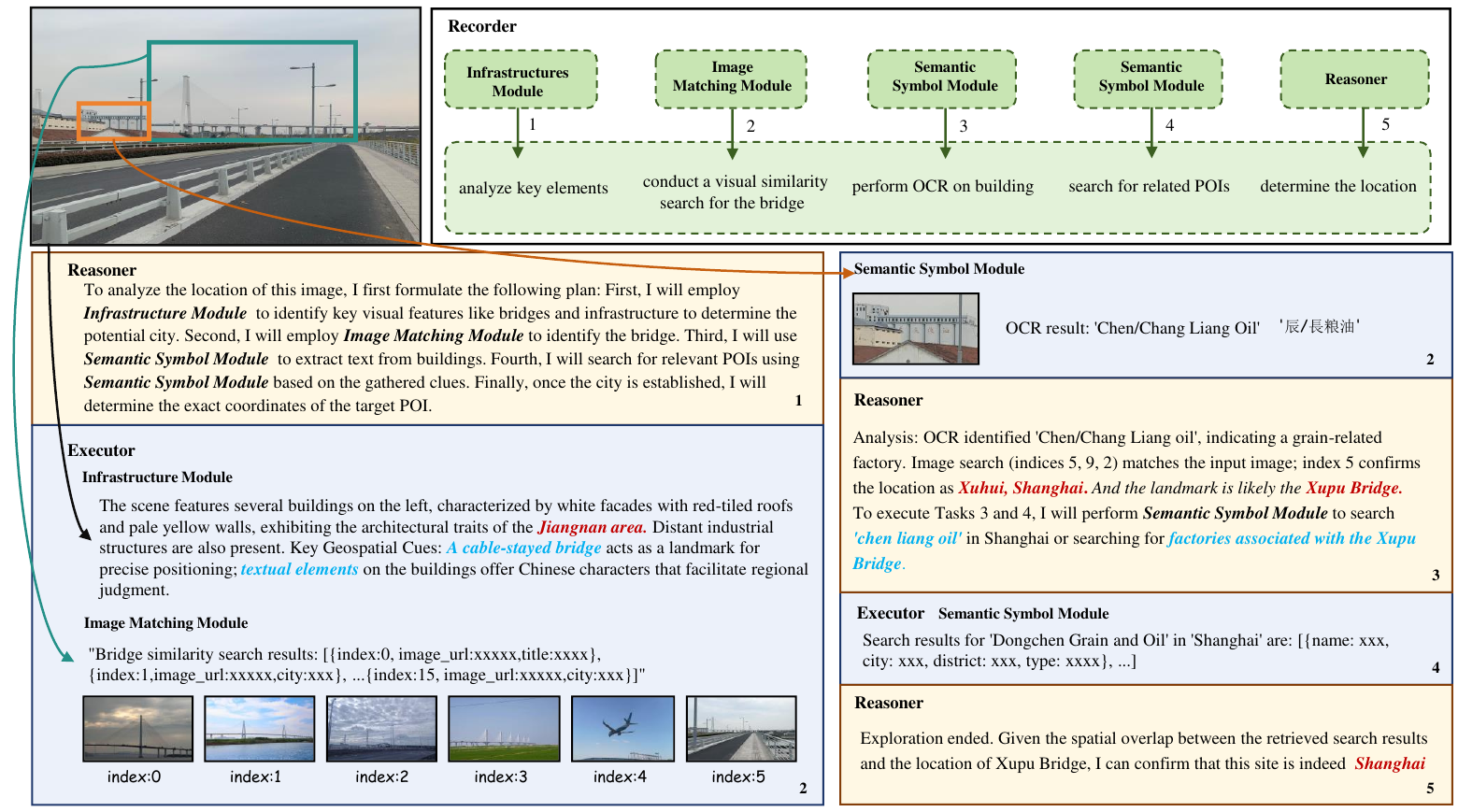}  % 核心命令：width设置宽度
    \caption{Complete localization reasoning trajectory of LocationAgent on an input image.}  % 图片标题
    \label{fig:4}  % 图片标签（用于交叉引用）
\end{figure*}

\begin{table}[t]
\centering
\caption{Localization accuracy (\%) of different methods on CCL-Bench.}
\label{tab:1}
\resizebox{\columnwidth}{!}{
\begin{tabular}{lrrrrr}
\toprule
Method & 1km & 25km & 200km & 750km & 2500km \\
\midrule
\multicolumn{6}{l}{\cellcolor[HTML]{EFEFEF}\bfseries\small Implicit Alignment} \\
GeoCLIP   & 7.00  & 16.33 & 28.33 & 50.33 & 91.33  \\
RFM-YFCC  & 7.33  & 20.33 & 28.33 & 49.33 & 89.00  \\
\addlinespace
\multicolumn{6}{l}{\cellcolor[HTML]{EFEFEF}\bfseries\small Explicit Alignment (fine-tuned reasoning models)} \\
GeoReasoner-7B & 0.33 & 13.33 & 29.33 & 54.67 & 93.00 \\
GLOBE-7B       & 3.67 & 21.00 & 38.67 & 63.67 & 95.33  \\
\addlinespace
\multicolumn{6}{l}{\cellcolor[HTML]{EFEFEF}\bfseries\small Explicit Alignment (closed-source reasoning models)} \\
Seed-1.6-Vision & 2.00 & 47.67 & 64.33 & 84.33 & 99.67  \\
GPT-5           & 7.67 & 40.33 & 55.33 & 81.00 & 99.67 \\
Gemini-2.5-pro  & 8.67 & 39.33 & 53.67 & 75.33 & 99.00  \\
\addlinespace
Ours            & 52.33 & 82.00 & 100.00 & 100.00 & 100.00 \\
\bottomrule
\end{tabular}
}
\end{table}

\subsection{Image Geolocation Accuracy Analysis}
We compare location accuracy across a range of methods, including implicit alignment approaches such as GeoCLIP\citep{vivanco2023geoclip} and RFM-YFCC\citep{dufour2025around}, training-based explicit alignment methods such as GeoReasoner\citep{Li2024georeansoner} and GLOBE\citep{li2025globe}, as well as advanced closed-source reasoning models.

Following standard evaluation protocols, we measure the percentage of predictions within predefined distance thresholds from the ground truth. Additionally, to reflect CCL-Bench’s regional focus, we introduce city level accuracy as a complementary metric.
\subsubsection{Distance-Threshold Accuracy Analysis}
As shown in Table\ref{tab:1} , implicit methods are consistently outperformed by explicit reasoning-based approaches. Within the explicit category, fine-tuned small models lag behind closed-source large reasoning models: at the 25km threshold, closed-source models achieve around 40\% accuracy, whereas fine-tuned small models reach only 13\% and 21\%.

Although these small models report comparable 25 km accuracies (40\%–60\%) on public benchmarks such as Im2GPS3K and MP16-Reason, their performance drops sharply on CCL-Bench. This discrepancy indicates substantial overlap between their fine-tuning data and existing benchmarks, leading to strong indomain performance but poor generalization to the real world distributions represented in CCL-Bench.

In contrast, our method consistently outperforms all baselines across all scales, achieving over 50\% accuracy at the stringent 1 km threshold and reaching 100\% accuracy beyond 200 km. These results demonstrate superior generalization and validate the effectiveness of integrating external tools for MLLM-based geographic reasoning.

\begin{table}[t]
\centering
\caption{City-level accuracy (\%) and Location Compliance (\%) of different methods on CCL-Bench.}
\label{tab:2}
\begin{tabular}{lrrr}
\toprule
Method
& \begin{tabular}[c]{@{}c@{}}ACC \\ City\end{tabular}
& \begin{tabular}[c]{@{}c@{}}ACC \\ Loglat\end{tabular}
& \begin{tabular}[c]{@{}c@{}}Location \\ Compliance\end{tabular} \\
\midrule
\multicolumn{4}{l}{\cellcolor[HTML]{EFEFEF}\bfseries\small Implicit Alignment} \\
GeoCLIP    & 14.33 & 14.33 & /     \\
RFM-YFCC   & 15.00 & 15.00 & /     \\
\addlinespace
\multicolumn{4}{l}{\cellcolor[HTML]{EFEFEF}\bfseries\small Explicit Alignment (fine-tuned reasoning models)} \\
GeoReasoner-7B & 18.33 & 17.67 & 78.33 \\
GLOBE-7B       & 26.33 & 23.33 & 60.30 \\
\addlinespace
\multicolumn{4}{l}{\cellcolor[HTML]{EFEFEF}\bfseries\small Explicit Alignment (closed-source reasoning models)} \\
Seed-1.6-Vision & 58.00 & 52.67 & 83.00 \\
GPT-5           & 45.33 & 44.67 & 94.00 \\
Gemini-2.5-pro  & 48.00 & 44.00 & 89.33 \\
\addlinespace
Ours & 84.67 & 100.00 & 100.00 \\
\bottomrule
\end{tabular}
\end{table}
\subsubsection{City-level Correctness Analysis}
Beyond distance-based metrics, we evaluate city level localization accuracy using two strategies. Results are reported in Table\ref{tab:2}. Direct prediction of city names and reverse geocoding of predicted coordinates. Explicit reasoning-based methods consistently outperform implicit ones, with closed-source large reasoning models surpassing fine-tuned small models. Our method further achieves the best overall performance. We also find that directly predicting city names generally yields higher accuracy than coordinate-to-city mapping.

To assess output coherence, we define the Location Compliance as the agreement between the predicted city name and that derived from the predicted coordinates. As shown in Table\ref{tab:2}, consistency varies widely across models, ranging from ~60\% for GLOBE to ~80\% for Seed, while GPT-5 achieves the highest rate at 95\%, indicating substantial gaps in current models’ forward and reverse geocoding reliability.

\subsection{Reasoning Trajectory Analysis}
Fig.\ref{fig:4} illustrates the workflow of LocationAgent during an image geolocation task. Unlike traditional single pass reasoning, LocationAgent exhibits pronounced characteristics of adaptive planning and parallel execution. 

At the initial stage, the Reasoner concurrently schedules the Executor’s text recognition and infrastructure modules, leveraging a hybrid “search and non-search” mode to rapidly identify specific structures (e.g., Jiangnan-style bridges) and relevant textual clues. Subsequently, through the state chain maintained by the Recorder, the Reasoner cross-references the initially retrieved bridge information with the recognized industrial buildings. This spatial constraint satisfaction logic effectively eliminates ambiguous candidate regions. Finally, after obtaining a set of POI candidates returned by the Executor, the system employs the semantic symbol module to determine the administrative division, ultimately converging the location precisely to Shanghai. This process demonstrates a rigorous inference flow from macro-level priors to micro-level evidence.

\begin{table}[t]
\centering
\caption{Ablation results of different search tools.}
\label{tab:3}
\begin{tabular}{lrrrr}
\toprule
Method & 1km & 25km & 200km & ACC-City \\
\midrule
LocationAgent              & 54.33 & 82.00 & 100.00 & 84.67 \\
w/o image search  & 21.33 & 43.67 & 99.00  & 49.67 \\
w/o text search   & 36.33 & 76.17 & 100.00 & 75.33 \\
w/o all tools     & 2.00  & 47.67 & 64.33  & 58.00 \\
\bottomrule
\end{tabular}
\end{table}

\subsection{Ablation Studies}
To investigate the impact of our design choices, we conduct ablation studies along two dimensions: (1) tool configuration and (2) base model configuration.
\subsubsection{Impact of Tool Configuration}
Within the LocationAgent framework, search tools are the primary means of acquiring external location information and include Image Search and Text Search. As shown in Table\ref{tab:3}, removing either tool results in a clear drop in localization accuracy, highlighting the importance of external evidence for high fidelity geolocation.

Overall, Image Search contributes more substantially to performance. This is consistent with the intuition that image-based retrieval operates directly on visual content, avoiding information loss introduced when visual clues are translated into text and thus preserving higher query–context consistency.

Notably, when only Text Search is retained, LocationAgent’s city level accuracy falls below that of the vanilla base model. We attribute this to our strict prompting strategy, which requires all predictions to be grounded in retrieved evidence and disallows unsupported guessing. Under Text Search alone, the model tends to issue coarse queries (e.g., “cities with supertall skyscrapers”), which often yield ambiguous results, sometimes misleading the agent compared to a base model that can rely on internal parametric knowledge.

\subsubsection{Impact of the Reasoning Model}
\begin{figure}[H] % htbp：浮动体位置参数（here/top/bottom/page）
    \centering  % 图片居中
    \includegraphics[width=0.4\textwidth]{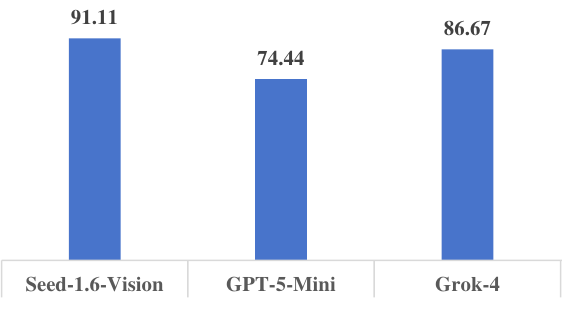}  % 核心命令：width设置宽度
    \caption{City-level accuracy (\%) of different base models under the LocationAgent framework.}  % 图片标题
    \label{fig:5}  % 图片标签（用于交叉引用）
\end{figure}
In addition to search tools, the reasoning model responsible for tool scheduling and strategic planning is a critical component of the LocationAgent architecture.Results are reported in Fig.\ref{fig:5} To evaluate its impact, we conduct ablation experiments by employing various closed-source reasoning models as the Reasoner, including GPT-5-Mini and Grok-4. 
The evaluation reveals a consistent performance trend: models with larger parameter scales generally outperform their smaller counterparts, within the LocationAgent framework. Among the evaluated candidates, Seed-1.6-Vision achieves the highest overall performance, surpassing Grok-4 under identical experimental conditions.

\section{Conclusion}
This paper proposes a new paradigm for image geolocation, in which complex geographic reasoning is decoupled from constrained parametric memory into two components—reasoning strategies and evidence verification—through the LocationAgent framework. To validate this approach, we construct CCL-Bench, which not only fills the gap in Chinese region data but also introduces challenging samples from real world internet environments, providing a rigorous testbed for evaluating expert level reasoning capabilities of geolocation models. Extensive experiments demonstrate that LocationAgent consistently outperforms state-of-the-art methods in both localization accuracy and generalization. Notably, LocationAgent achieves over 50\% accuracy at a stringent threshold, significantly narrowing the gap between machine performance and human-level expertise in complex geographic reasoning. In the future, we will continue to focus on optimizing agent architectures and expanding tool designs, further exploring the evolution of next generation intelligent maps and location-based services.

\bibliographystyle{named}
\bibliography{ijcai26}

\end{document}